\documentclass[10pt,twocolumn,letterpaper]{article}
\pdfoutput=1

\usepackage{iccv}
\usepackage{times}
\usepackage{epsfig}
\usepackage{graphicx}
\usepackage{amsmath}
\usepackage{amssymb}
\usepackage{booktabs}
\usepackage{multirow,array}
\usepackage{color}
\usepackage{bm}
\usepackage{subcaption}
\usepackage{tikz}
\usetikzlibrary{scopes,shadows,arrows,positioning}
\usetikzlibrary{decorations.markings}
\usetikzlibrary{shapes.symbols}
\usepackage{ifthen}
\usepackage[sort]{cite}			%
\usepackage[font=footnotesize,subrefformat=parens]{subcaption}
\usepackage{tabu}           %
\usepackage{xcolor}            %

\usepackage[pagebackref=true,breaklinks=true,letterpaper=true,colorlinks,bookmarks=false]{hyperref}

\iccvfinalcopy %

\def\iccvPaperID{} %

\ificcvfinal\fi

\definecolor{Plum}{RGB}{142, 69, 133}
\definecolor{Cyan}{RGB}{0, 255, 255} %
\definecolor{Red3}{HTML}{a40000}
\definecolor{Green3}{HTML}{4e9a06} %

\definecolor{tabutter}{rgb}{0.98824, 0.91373, 0.30980}		%
\definecolor{ta2butter}{rgb}{0.92941, 0.83137, 0}		%
\definecolor{ta3butter}{rgb}{0.76863, 0.62745, 0}		%

\definecolor{taorange}{rgb}{0.98824, 0.68627, 0.24314}		%
\definecolor{ta2orange}{rgb}{0.96078, 0.47451, 0}		%
\definecolor{ta3orange}{rgb}{0.80784, 0.36078, 0}		%

\definecolor{tachocolate}{rgb}{0.91373, 0.72549, 0.43137}	%
\definecolor{ta2chocolate}{rgb}{0.75686, 0.49020, 0.066667}	%
\definecolor{ta3chocolate}{rgb}{0.56078, 0.34902, 0.0078431}	%

\definecolor{tachameleon}{rgb}{0.54118, 0.88627, 0.20392}	%
\definecolor{ta2chameleon}{rgb}{0.45098, 0.82353, 0.086275}	%
\definecolor{ta3chameleon}{rgb}{0.30588, 0.60392, 0.023529}	%

\definecolor{taskyblue}{rgb}{0.44706, 0.56078, 0.81176}		%
\definecolor{ta2skyblue}{rgb}{0.20392, 0.39608, 0.64314}	%
\definecolor{ta3skyblue}{rgb}{0.12549, 0.29020, 0.52941}	%

\definecolor{taplum}{rgb}{0.67843, 0.49804, 0.65882}		%
\definecolor{ta2plum}{rgb}{0.45882, 0.31373, 0.48235}		%
\definecolor{ta3plum}{rgb}{0.36078, 0.20784, 0.4}		%

\definecolor{tascarletred}{rgb}{0.93725, 0.16078, 0.16078}	%
\definecolor{ta2scarletred}{rgb}{0.8, 0, 0}			%
\definecolor{ta3scarletred}{rgb}{0.64314, 0, 0}			%

\definecolor{taaluminium}{rgb}{0.93333, 0.93333, 0.92549}	%
\definecolor{ta2aluminium}{rgb}{0.82745, 0.84314, 0.81176}	%
\definecolor{ta3aluminium}{rgb}{0.72941, 0.74118, 0.71373}	%

\definecolor{tagray}{rgb}{0.53333, 0.54118, 0.52157}		%
\definecolor{ta2gray}{rgb}{0.33333, 0.34118, 0.32549}		%
\definecolor{ta3gray}{rgb}{0.18039, 0.20392, 0.21176}		%

\newcommand{\reffig}[1]{Figure~\ref{fig:#1}}
\newcommand{\refsec}[1]{Section~\ref{sec:#1}}

\newcommand{\reftbl}[1]{Table~\ref{tbl:#1}}

\newcommand{\lblfig}[1]{\label{fig:#1}}
\newcommand{\lblsec}[1]{\label{sec:#1}}
\newcommand{\lbleq}[1]{\label{eq:#1}}
\newcommand{\lbltbl}[1]{\label{tbl:#1}}

\newcommand{\rot}{\rotatebox{90}}
\definecolor{lightpink}{rgb}{1.0, 0.71, 0.76}
\definecolor{wildwatermelon}{rgb}{0.99, 0.42, 0.52}

\newcommand\blfootnote[1]{%
\begingroup
\renewcommand\thefootnote{}\footnote{#1}%
\addtocounter{footnote}{-1}%
\endgroup
}

\begin{document}

\title{Constrained Convolutional Neural Networks for\\
Weakly Supervised Segmentation}		%

\author{Deepak Pathak\qquad Philipp Kr\"ahenb\"uhl \qquad Trevor Darrell\\
University of California, Berkeley\\
{\tt\small \{pathak,philkr,trevor\}@cs.berkeley.edu}
}

\maketitle
\begin{abstract}
We present an approach to learn a dense pixel-wise labeling from image-level tags.
Each image-level tag imposes constraints on the output labeling of a Convolutional Neural Network (CNN) classifier.
We propose Constrained CNN (CCNN), a method which uses a novel loss function to optimize for any set of linear constraints on the output space (i.e. predicted label distribution) of a CNN.
Our loss formulation is easy to optimize and can be incorporated directly into standard stochastic gradient descent optimization.
The key idea is to phrase the training objective as a biconvex optimization for linear models, which we then relax to nonlinear deep networks.
Extensive experiments demonstrate the generality of our new learning framework.
The constrained loss yields state-of-the-art results on weakly supervised semantic image segmentation.
We further demonstrate that adding slightly more supervision can greatly improve the performance of the learning algorithm.
\end{abstract}

\blfootnote{The implementation code and trained models are available at the author's website.}

\section{Introduction}
\label{sec:intro}
In recent years, standard computer vision tasks, such as recognition or classification, have made tremendous progress.
This is primarily due to the widespread adoption of Convolutional Neural Networks (CNNs)~\cite{fukushima1980neocognitron,krizhevsky2012imagenet,lecun1989backpropagation}.
Existing models excel by their capacity to take advantage of massive amounts of fully supervised training data~\cite{imagenet}.
This reliance on full supervision is a major limitation on scalability with respect to the number of classes or tasks.
For structured prediction problems, such as semantic segmentation, fully supervised, i.e. pixel-level, labels are both expensive and time consuming to obtain.
Summarization of the semantic-labels in terms of weak supervision, e.g. image-level tags or bounding box annotations, is often less costly.
Leveraging the full potential of these weak annotations is challenging, and existing approaches are susceptible to diverging into bad local optima from which recovery is difficult~\cite{Hoffman_CVPR2015,pathak2014fully,verbeek2014_cvpr}.

In this paper, we present a framework to incorporate weak supervision into the learning procedure through a series of linear constraints.
In general, it is easier to express simple constraints on the output space than to craft regularizers or adhoc training procedures to guide the learning.
In semantic segmentation, such constraints can describe the existence and expected distribution of labels from image-level tags.
For example, given a car is present in an image, a certain number of pixels should be labeled as car.

\begin{figure}[t]
\begin{center}
  \begin{tikzpicture}[auto,tag/.style={rectangle, draw=black, thick, text width=5em, align=center, rounded corners, minimum height=2em}]
    \node[inner sep=0,label={Input Image}] (image) at ( 0,0) {\includegraphics[width=0.4\linewidth]{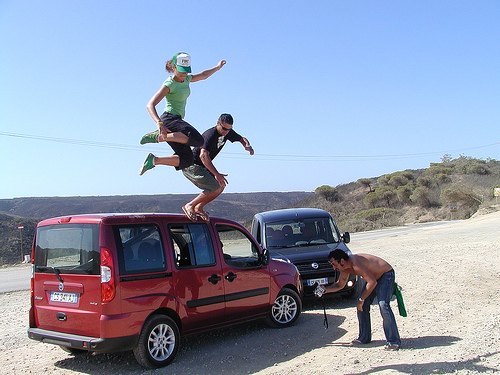}};
    \node[inner sep=0,label={Predicted labeling}] (output) at (4.5,0) {\includegraphics[width=0.4\linewidth]{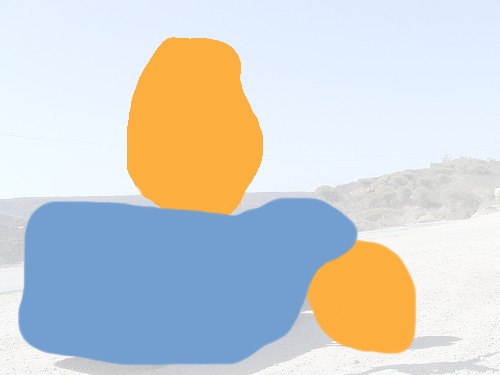}};
    \node[font=\footnotesize](output_l1) at (4.1, 0.5) {Person};
    \node[font=\footnotesize](output_l2) at (4,-0.6) {Car};
    \node[tag, draw=black!80, fill=black!10, minimum height=6em, minimum width=9em] (label) at (0,-2.75) {};
    \node(cr) at (0,-4.1) {Weak Labels};
    \node[tag, fill=taorange] (l1) at (0,-2.25) {\bf Person};
    \node[tag, fill=taskyblue] (l2) at (0,-3.25) {\bf Car};
    \draw[-{stealth},shorten >=1mm,shorten <=1mm,ultra thick] (image)--node[above]{CNN}(output);
    \fill [gray!30] (2.7,-4.)--(3.2,-2.6)--(3.3,-2.7)--(2.8,-4.1)--cycle;
    \fill [gray!30] (3,-2.8)--(4.75,-2.2)--(4.80,-2.3)--(3.05,-2.9)--cycle;
    \fill [gray!30] (4.45,-2.1)--(5.7,-3.4)--(5.65,-3.53)--(4.4,-2.23)--cycle;
    \draw[draw=ta2gray,ultra thick] (2.7,-4.)--(3.2,-2.6);
    \draw[draw=ta2gray,ultra thick] (3,-2.8)--(4.75,-2.2);
    \draw[draw=ta2gray,ultra thick] (4.45,-2.1)--(5.7,-3.4);
    \node(cr) at (4.5,-4.1) {Constrained Region};
    \draw[-{stealth},shorten >=1mm,shorten <=1mm,ultra thick] (label)--(2.8,-2.75);
    \node[circle, draw=black, fill=ta3scarletred, ultra thick] (no) at (3.6,-1.75) {};
    \node[circle, draw=black, fill=ta3chameleon, ultra thick] (ld) at (4.1,-2.75) {};
    \draw[-{latex},ultra thick] (no)--(ld);
    \node[text width=1.5cm,font=\footnotesize](cr) at (3,-1.75) {Network \mbox{Output}};
    \node[text width=1.5cm,align=center,font=\footnotesize](cr) at (4.1,-3.3) {Latent \mbox{Distribution}};
    \draw[-{latex},ultra thick] (output)--(no);
    \draw[-{latex},ultra thick, dashed] (output)--(ld);
  \end{tikzpicture}
\end{center}
   \caption{We train convolutional neural networks from a set of linear constraints on the output variables.
   The network output is encouraged to follow a latent probability distribution, which lies in the constraint manifold.
   The resulting loss is easy to optimize and can incorporate arbitrary linear constraints.}
\lblfig{teaser}
\end{figure}
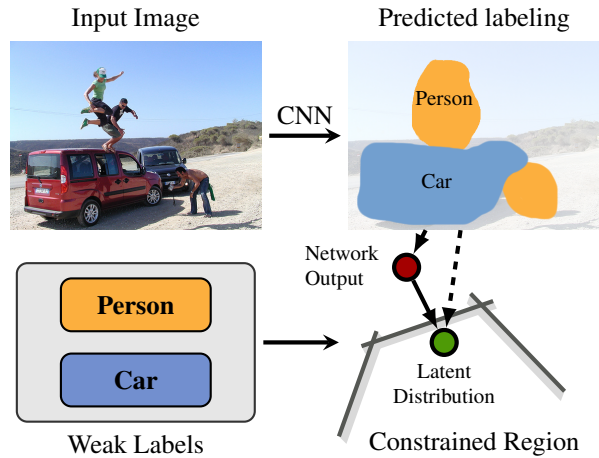

We propose Constrained CNN (CCNN), a method which uses a novel loss function to optimize convolutional networks with arbitrary linear constraints on the structured output space of pixel labels.
The non-convex nature of deep nets makes a direct optimization of the constraints difficult.
Our key insight is to model a distribution over latent ``ground truth'' labels while the output of the deep net follows this latent distribution as closely as possible.
This allows us to enforce the constraints on the latent distribution instead of the network output, which greatly simplifies the resulting optimization problem.
The resulting objective is a biconvex problem for linear models.
For deep nonlinear models, it results in an alternating convex and gradient based optimization which can be naturally integrated into standard stochastic gradient descent (SGD).
As illustrated in \reffig{teaser}, after each iteration the output is pulled towards the closest point on the constrained manifold of plausible semantic segmentation.
Our Constrained CNN is guided by weak annotations and trained end-to-end.

We evaluate CCNN on the problem of multi-class semantic segmentation with varying levels of weak supervision defined by different linear constraints.
Our approach achieves state-of-the-art performance on Pascal VOC 2012 compared to other weak learning approaches.
It does not require pixel-level labels for any objects during the training time, but infers them directly from the image-level tags.
We show that our constrained optimization framework can incorporate additional forms of weak supervision, such as a rough estimate of the size of an object.
The proposed technique is general, and can incorporate many forms of weak supervision.

\section{Related Work}
\label{sec:related}
Weakly supervised learning seeks to capture the signal that is common to all the positives but absent from all the negatives.
This is challenging due to nuisance variables such as pose, occlusion, and intra-class variation.
Learning with weak labels is often phrased as Multiple Instance Learning~\cite{dietterich1997solving}.
It is most frequently formulated as a maximum margin problem, although boosting~\cite{ali_cvpr14,zhang2005multiple} and Noisy-OR models~\cite{heckerman2013tractable} have been explored as well.
The multiple instance max-margin classification problem is non-convex and solved as an alternating minimization of a biconvex objective~\cite{andrews2002_nips}.
MI-SVM~\cite{andrews2002_nips} or LSVM~\cite{felzenszwalb2010_pami} are two classic methods in this paradigm.
This setting naturally applies to weakly-labeled detection~\cite{wang_eccv14,song2014_icml}.
However, most of these approaches are sensitive to the initialization of the detector~\cite{verbeek2014_cvpr}.
Several heuristics have been proposed to address these issues~\cite{singh2012unsupervised,song2014_icml}, however they are usually specific to detection.

Traditionally, the problem of weak segmentation and scene parsing with image level labels has been addressed using graphical models, and parametric structured models~\cite{vezhnevets2010towards,vezhnevets2012weakly,zhang2014representative}.
Most works exploit low-level image information to connect regions similar in appearance~\cite{vezhnevets2010towards}.
Chen \etal~\cite{chen2014enriching} exploit top-down segmentation priors based on visual subcategories for object discovery.
Pinheiro \etal~\cite{pinheiro2014weakly} and Pathak \etal~\cite{pathak2014fully} extend the multiple-instance learning framework from detection to semantic segmentation using CNNs.
Their methods iteratively reinforce well-predicted outputs while suppressing erroneous segmentations contradicting image-level tags.
Both algorithms are very sensitive to the initialization, and rely on carefully pretrained classifiers for all layers in the convolutional network.
In contrast, our constrained optimization is much less sensitive and recovers a good solution from any random initialization of the classification layer.

Papandreou \etal~\cite{papandreou2015weakly} include an adaptive bias into the multi-instance learning framework.
Their algorithm boosts classes known to be present and suppresses all others.
We show that this simple heuristic can be viewed as a special case of a constrained optimization, where the adaptive bias controls the constraint satisfaction.
However the constraints that can be modeled by this adaptive bias are limited and cannot leverage the full power of weak labels.
In this paper, we show how to apply  more general linear constraints which lead to better segmentation performance.
Constrained optimization problems have long been approximated by artificial neural networks~\cite{zak1994solving}.
These models are usually non-parametric, and solve just a single instance of a linear program.
Platt \etal~\cite{platt1988constrained} show how to optimize equality constraints on the output of a neural network.
However the resulting objective is highly non-convex, which makes a direct minimization hard.
In this paper, we show how to optimize a constrained objective by alternating between a convex and gradient-based optimization.

The resulting algorithm is similar to generalized expectation~\cite{mann2010generalized} and posterior regularization~\cite{ganchev2010posterior} in natural language processing.
Both methods train a parametric model that matches certain expectation constraints by applying a penalty to the objective function.
Generalized expectation adds the expected constraint penalty directly to objective, which for convolutional networks is hard and expensive to evaluate directly.
Ganchev~\etal~\cite{ganchev2010posterior} constrain an auxiliary variable yielding an algorithm similar to our objective in dual space.

\section{Preliminaries}
\label{sec:preliminaries}
We define a pixel-wise labeling for an image $I$ as a set of random variables $X=\{x_0,\ldots,x_n\}$ where $n$ is the number of pixles in an image.
$x_i \in \mathcal{L}$ takes one of $m$ discrete labels $\mathcal{L}=\{1,\ldots,m\}$.
CNN models a probability distribution $Q(X|\theta,I)$ over those random variables, where $\theta$ are the parameters of the network.
The distribution is commonly modeled as a product of independent marginals $Q(X|\theta,I) = \prod_i q_i(x_i|\theta,I)$, where each of the marginal represents a softmax probability:

\begin{align} \label{eq:softmax}
q_i(x_i|\theta,I) = \frac{1}{Z_i} \exp \big(f_i(x_i;\theta,I)\big)
\end{align}
where $Z_i = \sum_{l\in\mathcal{L}} \exp  \big(f_i(l;\theta,I)\big)$ is the partition function of a pixel $i$. 
The function $f_i$ represents the real-valued score of the neural network.
A higher score corresponds to a higher likelihood.

Standard learning algorithms aim to maximize the likelihood of the observed training data under the model.
This requires full knowledge of the ground truth labeling, which is not available in the weakly supervised setting.
In the next section, we show how to optimize the parameters of a CNN using some high-level constraints on the distribution of output labeling.
An overview of this is given in \reffig{overview}.
In \refsec{constraints}, we then present a few examples of useful constraints for weak labeling.

\section{Constrained Optimization}
\label{sec:method}

\begin{figure}[t]
\begin{center}
\includegraphics[width=\linewidth]{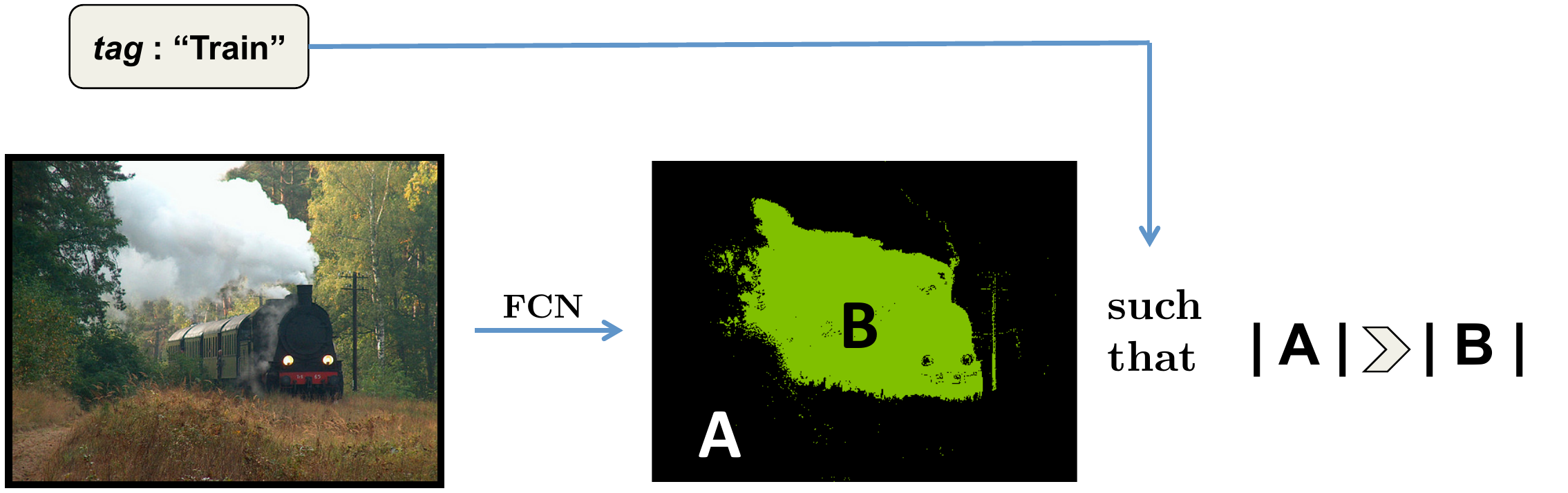}
\end{center}
   \caption{Overview of our weak learning pipeline.
   Input image is passed through a fully convolutional network (FCN) which produces an output labeling.
   The model is trained such that the output labeling follows a set of simple linear constraints imposed by image level tags.}
\lblfig{overview}
\end{figure}

For notational convenience, let $\vec{Q}_I$ be the vectorized form of network output $Q(X|\theta,I)$.
The Constrained CNN (CCNN) optimization can be framed as:
\begin{align}
\text{find} & & &\theta & \nonumber \\
\text{subject to} & & & A_I \vec{Q}_I \geq \vec b_I  \qquad \forall I \lbleq{objective}, &
\end{align}
where $A_I \in \mathbb{R}^{k \times nm}$ and $\vec b_I \in \mathbb{R}^k$ enforce $k$ individual linear constraints on the output distribution of the convnet on image $I$.
In theory, many outputs $\vec{Q}_I$ satisfy these constraints.
However all network outputs are parametrized by a single parameter vector $\theta$, which ties the output space of different $\vec{Q}_I$ together.
In practice, this leads to an output that is both consistent with the input image and the constraints imposed by the weak labels.

For notational simplicity, we derive our inference algorithm for a single image with $A = A_I$, $\vec b = \vec b_I$ and $\vec{Q} = \vec{Q}_I$.
The entire derivation generalizes to an arbitrary number of images and constraints.
Constraints include for example lower and upper bounds on the expected number of foreground and background pixel labels in a scene.
For more examples, see \refsec{constraints}.
In the first part of this section, we assume that all constraints are satisfiable, meaning there always exists a parameter setting $\theta$ such that $A\vec Q \ge \vec b$.
In \refsec{slack}, we lift this assumption by adding slack variables to each of the constraints.

While problem \eqref{eq:objective} is convex in the network output $Q$, it is generally not convex with respect to the network parameters $\theta$.
For any non-linear function $Q$, the matrix $A$ can be chosen such that the constraint is an upper or lower bound to $Q$, one of which is non-convex.
This makes a direct optimization hard.
As a matter of fact, not even log-linear models, such as logistic regression, can be directly optimized under this objective.
Alternatively, one could optimize the Lagrangian dual of \eqref{eq:objective}.
However this is computationally very expensive, as we would need to optimize an entire convolutional neural network in an inner loop of a dual descent algorithm.

In order to efficiently optimize problem \eqref{eq:objective}, we introduce a latent probability distribution $P(X)$ over the semantic labels $X$.
We constrain $P(X)$ to lie in the feasibility region of the constrained objective while removing the constraints on the network output $Q$.
We then encourage $P$ and $Q$ to model the same probability distribution by minimizing their respective KL-divergence.
The resulting problem is defined as
\begin{align}
\underset{\theta,P}{\text{minimize}} & & & D\left(P(X)\|Q(X|\theta)\right) \nonumber & \\
\text{subject to} & & & A \vec{P} \geq \vec b, \qquad \sum_X P(X)=1, &\lbleq{KL}
\end{align}
where $D\left(P(X)\|Q(X|\theta)\right) = \sum_{X} P(X)\log P(X) - \mathbb{E}_{X\sim P}\left[\log Q(X|\theta)\right]$
and $\vec P$ is the vectorized version of $P(X)$.
If the constraints in \eqref{eq:objective} are satisfiable then the problems \eqref{eq:objective} and \eqref{eq:KL} are equivalent with a solution of \eqref{eq:KL} at $P$ that is equal to the feasible $Q$.
This equality implies that $P(X)$ can be modeled as a product of independent marginals $P(X) = \prod_i p_i(x_i)$ without loss of generality, with a minimum at $p_i(x_i) = q_i(x_i|\theta)$.
A detailed proof is provided in the supplementary material.

The new objective is much easier to optimize, as it decouples the constraints from the network output.
For fixed network parameters $\theta$, the problem is convex in $P$.
For a fixed latent distribution $P$, the problem reduces to a standard cross entropy loss which is optimized using stochastic gradient descent.

In the remainder of this section, we derive an algorithm to optimize problem \eqref{eq:KL} using block coordinate descent.
\refsec{optimization} solves the constrained optimization for $P$ while keeping the network parameters $\theta$ fixed.
\refsec{sgd} then incorporates this optimization into standard stochastic gradient descent, keeping $P$ fixed while optimizing for $\theta$.
Each step is guaranteed to decrease the overall energy of problem \eqref{eq:KL}, converging to a good local optimum.
At the end of this section, we show how to handle constraints that are not directly satisfiable by adding a slack variable to the loss.

\subsection{Latent distribution optimization}
\lblsec{optimization}
We first show how to optimize problem~\eqref{eq:KL} with respect to $P$ while keeping the convnet output fixed.
The objective function is convex with linear constraints, which implies Slaters condition and hence strong duality holds as long as the constraints are satisfiable \cite{boyd2004convex}.
We can therefore optimize problem~\eqref{eq:KL} by maximizing its dual function, i.e.,
\begin{equation}
 \mathcal{L}(\lambda) = \lambda^\top \vec{b} -\sum_{i=1}^{n} \log \sum_{l\in \mathcal{L}} \exp\big(f_i(l;\theta)+A_{i;l}^\top \lambda \big),\lbleq{dual}
\end{equation}
where $\lambda \ge 0$ are the dual variables pertaining to the inequality constraints and $f_i(l;\theta)$ is the score of the convnet classifier for pixel $i$ and label $l$.
$A_{i;l}$ is the column of $A$ corresponding to $p_i(l)$.
A detailed derivation of this dual function is provided in the supplementary material.

The dual function is concave and can be optimized globally using projected gradient ascent \cite{boyd2004convex}.
The gradient of the dual function is given by $\frac{\partial}{\partial\lambda} \mathcal{L}(\lambda) = \vec b - A \vec P$, which results into
$$
 p_i(x_i) = \frac{1}{Z_i} \exp\big(f_i(x_i;\theta)+A_{i;x_i}^\top \lambda \big),
$$
where $Z_i = \sum_l \exp\big(f_i(l;\theta)+A_{i;l}^\top \lambda \big)$ is the local partition function ensuring that the distribution $p_i(x_i)$ sums to one for $\forall x_i\in \mathcal{L}$.
Intuitively, the projected gradient descent algorithm increases the dual variables for all constraints that are not satisfied.
Those dual variables in turn adjust the distribution $p_i$ to fulfill the constraints.
The projected dual gradient descent algorithm usually converges within fewer than $50$ iterations, making the optimization highly efficient.

\begin{figure}[t]
\begin{center}
\includegraphics[width=\linewidth]{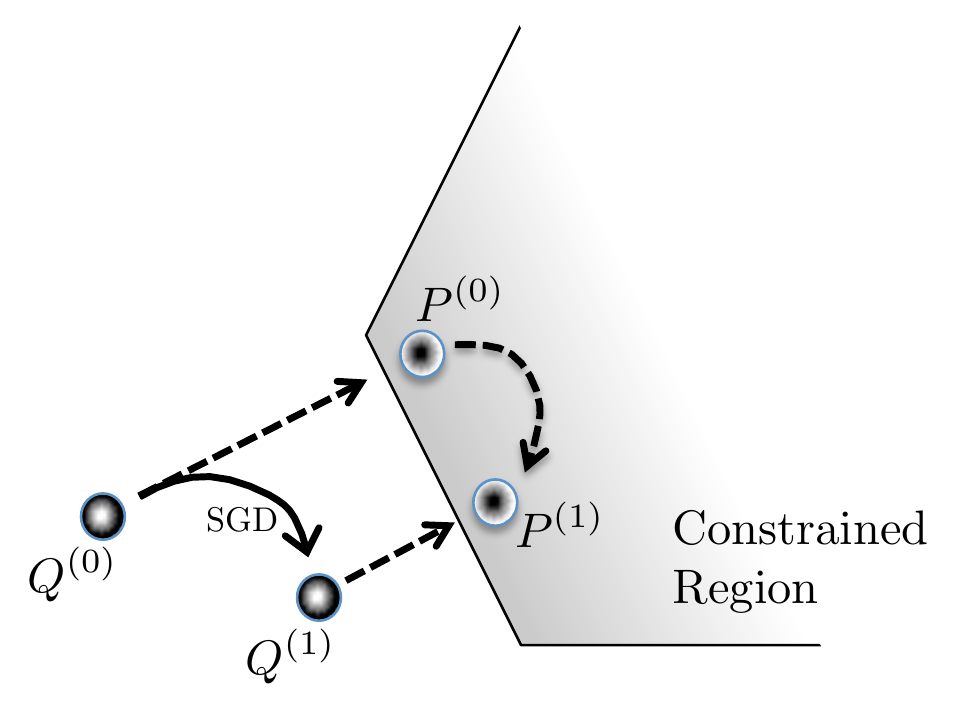}
\end{center}
   \caption{Illustration of our alternating convex optimization and gradient descent optimization for $t=0$.
   At each iteration $t$, we compute a latent probability distribution $P^{(t)}$ as the closest point in the constrained region.
   We then update the convnet parameters to follow $P^{(t)}$ as closely as possible using Stochastic Gradient Descent (SGD), which takes the convnet output from $Q^{(t)}$ to $Q^{(t+1)}$.}
\lblfig{sgd}
\end{figure}

Next, we show how to incorporate this estimate of $P(X)$ into the standard stochastic gradient descent algorithm.

\subsection{SGD}
\lblsec{sgd}

For a fixed latent distribution $P$, problem \eqref{eq:KL} reduces to the standard cross entropy loss
\begin{align}
 L(\theta) = -\sum_i \sum_{x_i} p_i(x_i) \log q_i(x_i|\theta) \lbleq{c_ent}.
\end{align}
The gradient of this loss function is given by \mbox{$\frac{\partial}{\partial \vec f_i(x_i)} L(\theta) = \vec q_i(x_i|\theta) - \vec p_i(x_i)$.}
For linear models, the loss function \eqref{eq:c_ent} is convex and can be optimized using any gradient based optimization.
For multi-layer deep networks, we optimize it using back-propagation and stochastic gradient descent (SGD) with momentum, as implemented in Caffe \cite{caffe}.

Theoretically, we would need to keep the latent distribution $P$ fixed for a few iterations of SGD until the objective value decreases.
Otherwise, we are not strictly guaranteed that the overall objective \eqref{eq:KL} decreases.
However, in practice we found inferring a new latent distribution at every step of SGD does not hurt the performance and leads to a faster convergence.

In summary, we optimize problem \eqref{eq:KL} using SGD, where at each iteration we infer a latent distribution $P$ which defines both our loss and loss gradient.
\reffig{sgd} shows an overview of the training procedure.
For more details, see \refsec{implementation}.

Up to this point, we assumed that all the constraints are simultaneously satisfiable.
While this might hold for carefully chosen constraints, our optimization should be robust to arbitrary linear constraints.
In the next section, we relax this assumption by adding a slack variable to the constraint set and show that this slack variable can then be easily integrated into the optimization.

\subsection{Constraints with slack variable}
\lblsec{slack}
We relax problem \eqref{eq:KL} by adding a slack $\xi\in\mathbb{R}^k$ to the linear constraints.
The slack is regularized using a hinge loss with weight $\beta\in \mathbb{R}^k$.
It results into the following optimization:
\begin{align}
\underset{\theta,P,\xi}{\text{minimize}} & & & D\left(P(X)\|Q(X|\theta)\right) + \beta^T\xi \nonumber &\\
\text{subject to} & & & A\vec{P} \geq \vec b\!-\!\xi, \quad \sum_{X}\!P(X)\!=\!1, \quad\xi\geq 0.  &\lbleq{slack}
\end{align}
This objective is now guaranteed to be satisfiable for any assignment to $P$ and any linear constraint.
Similar to~\eqref{eq:dual}, this is optimized using projected dual coordinate ascent.
The dual objective function is exactly same as~\eqref{eq:dual}.
The weighting term of the hinge loss $\beta$ merely acts as an upper bound on the dual variable i.e. $0\leq \lambda \leq \beta$. 
A detailed derivation of this loss is given in the supplementary material.

This slack relaxed loss allows the optimization to ignore certain constraints if they become too hard to enforce.
It also trades off between various competing constraints.

\section{Constraints for Weak Semantic Segmentation}
\label{sec:constraints}
We now describe all constraints we use for our weakly supervised semantic segmentation.
For each training image $I$, we are given a set of image-level labels $\mathcal{L}_I$.
Our constraints affect different parts of the output space depending on the image-level labels.
All the constraints are complementary, and each constraint exploits the set of image-level labels differently.

\paragraph{Suppression constraint}
The most natural constraint is to suppress any label $l$ that does not appear in the image.
\begin{align}
\sum_{i=1}^n p_i(l) \le 0 \qquad \forall \; l \notin \mathcal{L}_I. \lbleq{suppConstraint} &
\end{align}
This constraint alone is not sufficient, as a solution involving all background labels satisfies it perfectly.
We can easily address this by adding a lower-bound constraint for labels present in an image.

\paragraph{Foreground constraint}
\begin{align}
a_l \le \sum_{i=1}^n p_i(l) \qquad \forall \; l \in \mathcal{L}_I. \lbleq{foreConstraint} &
\end{align}
This foreground constraint is very similar to the commonly used multiple instance learning (MIL) paradigm, where at least one pixel is constrained to be positive \cite{andrews2002_nips,Hoffman_CVPR2015,pinheiro2014weakly,pathak2014fully}.
Unlike MIL, our foreground constraint can encourage multiple pixels to take a specific foreground label by increasing $a_l$.
In practice, we set $a_l = 0.05 n$ with a slack of $\beta=2$, where $n$ is the number of outputs of our network.

While this foreground constraint encourages some of the pixels to take a specific label, it is often not strong enough to encourage all pixels within an object to take the correct label.
We could increase $a_l$ to encourage more foreground labels, but this would over-emphasize small objects.
A more natural solution is to constrain the total number of foreground labels in the output, which is equivalent to constraining the overall area of the background label.

\begin{table*}
\begin{center}
\resizebox{\textwidth}{!}{%
\begin{tabular}{lc@{\hskip 2.5mm}c@{\hskip 2.5mm}c@{\hskip 2.5mm}c@{\hskip 2.5mm}c@{\hskip 2.5mm}c@{\hskip 2.5mm}c@{\hskip 2.5mm}c@{\hskip 2.5mm}c@{\hskip 2.5mm}c@{\hskip 2.5mm}c@{\hskip 2.5mm}c@{\hskip 2.5mm}c@{\hskip 2.5mm}c@{\hskip 2.5mm}c@{\hskip 2.5mm}c@{\hskip 2.5mm}c@{\hskip 2.5mm}c@{\hskip 2.5mm}c@{\hskip 2.5mm}c@{\hskip 2.5mm}cc}
\toprule
Method & \rot{bgnd} & \rot{aero} & \rot{bike} & \rot{bird} & \rot{boat} & \rot{bottle} & \rot{bus} & \rot{car} & \rot{cat} & \rot{chair} & \rot{cow} & \rot{table} & \rot{dog} & \rot{horse} & \rot{mbike} & \rot{person} & \rot{plant} & \rot{sheep} & \rot{sofa} & \rot{train} & \rot{tv} & mIoU \\
\midrule
MIL-FCN~\cite{pathak2014fully} & - & - & - & - & - & - & - & - & - & - & - & - & - & - & - & - & - & - & - & - & - & 24.9 \\
MIL-Base~\cite{pinheiro2014weakly} & 37.0 & 10.4 & 12.4 & 10.8 & 05.3 & 05.7 & 25.2 & 21.1 & 25.2 & 04.8 & 21.5 & 08.6 & 29.1 & 25.1 & 23.6 & 25.5 & 12.0 & 28.4 & 08.9 & 22.0 & 11.6 & 17.8 \\
MIL-Base w/ ILP~\cite{pinheiro2014weakly} & \textbf{73.2} & 25.4 & \textbf{18.2} & 22.7 & 21.5 & 28.6 & 39.5 & 44.7 & 46.6 & 11.9 & \textbf{40.4} & 11.8 & \textbf{45.6} & \textbf{40.1} & 35.5 & 35.2 & 20.8 & \textbf{41.7} & 17.0 & 34.7 & 30.4 & 32.6 \\
EM-Adapt w/o CRF~\cite{papandreou2015weakly} & 65.3 & 28.2 & 16.9 & 27.4 & 21.1 & 28.1 & 45.4 & 40.5 & 42.3 & 13.2 & 32.1 & 23.3 & 38.7 & 32.0 & 39.9 & 31.3 & 22.7 & 34.2 & 22.8 & 37.0 & 30.0 & 32.0 \\
EM-Adapt~\cite{papandreou2015weakly} & 67.2 & \textbf{29.2} & 17.6 & \textbf{28.6} & \textbf{22.2} & 29.6 & \textbf{47.0} & 44.0 & 44.2 & 14.6 & 35.1 & \textbf{24.9} & 41.0 & 34.8 & 41.6 & 32.1 & 24.8 & 37.4 & \textbf{24.0} & 38.1 & 31.6 & 33.8\\
\midrule
CCNN w/o CRF & 66.3 & 24.6 & 17.2 & 24.3 & 19.5 & 34.4 & 45.6 & 44.3 & 44.7 & 14.4 & 33.8 & 21.4 & 40.8 & 31.6 & 42.8 & 39.1 & 28.8 & 33.2 & 21.5 & 37.4 & 34.4 & 33.3 \\
CCNN & 68.5 & 25.5 & 18.0 & 25.4 & 20.2 & \textbf{36.3} & 46.8 & \textbf{47.1} & \textbf{48.0} & \textbf{15.8} & 37.9 & 21.0 & 44.5 & 34.5 & \textbf{46.2} & \textbf{40.7} & \textbf{30.4} & 36.3 & 22.2 & \textbf{38.8} & \textbf{36.9} & \textbf{35.3} \\
\bottomrule
\end{tabular}}
\end{center}
\vspace{-1em}
\caption{Comparison of weakly supervised semantic segmentation methods on PASCAL VOC 2012 validation set.}
\lbltbl{weak_fair}
\end{table*}

\paragraph{Background constraint}
\begin{align}
a_0 \le \sum_{i=1}^n p_i(0) \le b_0. \lbleq{backConstraint} &
\end{align}
Here $l=0$ is assumed to be the background label.
We apply both a lower and upper bound on the background label.
This indirectly controls the minimum and maximum combined area of all foreground labels.
We found $a_0=0.3n$ and $b_0=0.7n$ to work well in practice.

The above constraints are all complementary and ensure that the final labeling follows the image-level labels $\mathcal{L}_I$ as closely as possible.
If we also have access to the rough size of an object, we can exploit this information during training.
In our experiments, we show that substantial gains can be made by simply knowing if a certain object class covers more or less than $10\%$ of the image.

\paragraph{Size constraint}
We exploit the size constraint in two ways: We boost all classes larger than $10\%$ of the image by setting $a_l=0.1n$.
We also put an upper bound constraint on the classes $l$ that are guaranteed to be small
\begin{align}
\sum_{i=1}^n p_i(l) \le b_l. \lbleq{sizeConstraint} &
\end{align}
In practice, a threshold $b_l < 0.01n$ works slightly better than a tight threshold.

The EM-Adapt algorithm of Papandreou~\etal~\cite{papandreou2015weakly} can be seen as a special case of a constrained optimization problem with just suppression and foreground constraints.
The adaptive bias parameters then correspond to the Lagrangian dual variables $\lambda$ of our constrained optimization.
However in the original algorithm of Papandreou~\etal, the constraints are not strictly enforced especially when some of them conflict.
In \refsec{experiments}, we show that a principled optimization of those constraints, CCNN, leads to a substantial increase in performance.

\section{Implementation Details}
\label{sec:implementation}

In this section, we discuss the overall pipeline of our algorithm applied for semantic image segmentation. 
We consider the weakly supervised setting i.e. only image-level labels are present during training. 
At test time, the task is to predict semantic segmentation mask for a given image.
 
\paragraph{Learning}
The CNN architecture used in our experiments is derived from VGG 16-layer network~\cite{simonyan2014very}.
It was pre-trained on Imagenet 1K class dataset, and achieved winning performance on ILSVRC14.
We cast the fully connected layers into convolutions in a similar fashion as suggested in~\cite{long2014fully}, and the last fc8 layer with 1K outputs is replaced by that containing 21 outputs corresponding to 20 object classes in Pascal VOC and background class.
The overall network stride of this fully convolutional network is 32s.
However, we observe that the slightly modified architecture with the denser 8s network stride proposed in~\cite{chen2014semantic} gives better results in the weakly supervised training.
Unlike~\cite{pathak2014fully,pinheiro2014weakly}, we do not learn any weights of the last layer from Imagenet.
Apart from the initial pre-training, all parameters are finetuned only on Pascal VOC.
We initialize the weights of the last layer with random Gaussian noise.

The FCN takes in arbitrarily sized images and produces coarse heatmaps corresponding to each class in the dataset.
We apply the convex constrained optimization on these coarse heatmaps, reducing the computational cost.
The network is trained using SGD with momentum.
We follow \cite{long2014fully} and train our models with a batch size of $1$, momentum of $0.99$ and an initial learning rate of $1e$-$6$.
We train for $60000$ iterations, which corresponds to roughly $5$ epochs.
The learning rate is decreased by a factor of $0.1$ every $20000$ iterations.
We found this setup to outperform a batch size of $20$ with momentum of $0.9$~\cite{chen2014semantic}.
The constrained optimization for single image takes less than $30$ ms on a CPU single core, and could be accelerated using a GPU.
The total training time is 8-9 hrs, comparable to~\cite{long2014fully,papandreou2015weakly}.

\begin{table*}
\begin{center}
\resizebox{\textwidth}{!}{%
\begin{tabular}{lc@{\hskip 2.5mm}c@{\hskip 2.5mm}c@{\hskip 2.5mm}c@{\hskip 2.5mm}c@{\hskip 2.5mm}c@{\hskip 2.5mm}c@{\hskip 2.5mm}c@{\hskip 2.5mm}c@{\hskip 2.5mm}c@{\hskip 2.5mm}c@{\hskip 2.5mm}c@{\hskip 2.5mm}c@{\hskip 2.5mm}c@{\hskip 2.5mm}c@{\hskip 2.5mm}c@{\hskip 2.5mm}c@{\hskip 2.5mm}c@{\hskip 2.5mm}c@{\hskip 2.5mm}cc}
\toprule
Method & \rot{aero} & \rot{bike} & \rot{bird} & \rot{boat} & \rot{bottle} & \rot{bus} & \rot{car} & \rot{cat} & \rot{chair} & \rot{cow} & \rot{table} & \rot{dog} & \rot{horse} & \rot{mbike} & \rot{person} & \rot{plant} & \rot{sheep} & \rot{sofa} & \rot{train} & \rot{tv} & mIoU \\
\midrule
\multicolumn{22}{l}{Fully Supervised:}\\
SDS~\cite{BharathECCV2014} & 63.3 & 25.7 & 63.0 & 39.8 & 59.2 & 70.9 & 61.4 & 54.9 & 16.8 & 45.0 & 48.2 & 50.5 & 51.0 & 57.7 & 63.3 & 31.8 & 58.7 & 31.2 & 55.7 & 48.5 & 51.6\\
FCN-8s~\cite{long2014fully} & 76.8 & 34.2 & 68.9 & 49.4 & 60.3 & 75.3 & 74.7 & 77.6 & 21.4 & 62.5 & 46.8 & 71.8 & 63.9 & 76.5 & 73.9 & 45.2 & 72.4 & 37.4 & 70.9 & 55.1 & 62.2 \\
TTIC Zoomout~\cite{mostajabi2014feedforward} & 81.9 & 35.1 & 78.2 & 57.4 & 56.5 & 80.5 & 74.0 & 79.8 & 22.4 & 69.6 & 53.7 & 74.0 & 76.0 & 76.6 & 68.8 & 44.3 & 70.2 & 40.2 & 68.9 & 55.3 & 64.4\\
DeepLab-CRF~\cite{chen2014semantic} & 78.4 & 33.1 & 78.2 & 55.6 & 65.3 & 81.3 & 75.5 & 78.6 & 25.3 & 69.2 & 52.7 & 75.2 & 69.0 & 79.1 & 77.6 & 54.7 & 78.3 & 45.1 & 73.3 & 56.2 & \textbf{66.4}\\
\midrule
\multicolumn{22}{l}{Weakly Supervised:}\\
CCNN w/ tags & 24.2 & 19.9 & 26.3 & 18.6 & 38.1 & 51.7 & 42.9 & 48.2 & 15.6 & 37.2 & 18.3 & 43.0 & 38.2 & 52.2 & 40.0 & 33.8 & 36.0 & 21.6 & 33.4 & 38.3 & 35.6 \\
CCNN w/ size & 36.7 & 23.6 & 47.1 & 30.2 & 40.6 & 59.5 & 54.3 & 51.9 & 15.9 & 43.3 & 34.8 & 48.2 & 42.5 & 59.2 & 43.1 & 35.5 & 45.2 & 31.4 & 46.2 & 42.2 & 43.3 \\
CCNN w/ size (CRF tuned) & 42.3 & 24.5 & 56.0 & 30.6 & 39.0 & 58.8 & 52.7 & 54.8 & 14.6 & 48.4 & 34.2 & 52.7 & 46.9 & 61.1 & 44.8 & 37.4 & 48.8 & 30.6 & 47.7 & 41.7 & \textbf{45.1} \\
\bottomrule
\end{tabular}}
\end{center}
\vspace{-1em}
\caption{Results on PASCAL VOC 2012 test. We compare our results to the fully supervised state-of-the-art methods.}
\lbltbl{weak_test}
\end{table*}

\vspace{-2.5mm}
\paragraph{Inference}
At inference time, we optionally apply a fully connected conditional random field model~\cite{kk-eifcc-11} to refine the final segmentation.
We used the default parameter provided by the authors for all our experiments.

\section{Experiments}
\label{sec:experiments}
We analyze and compare the performance of our constrained optimization for varying levels of supervision: image-level tags and additional supervision such as object size information.
The objective is to learn models to predict dense multi-class semantic segmentation i.e. pixel-wise labeling for any new image.
We use the provided supervision with few simple spatial constraints on the output, and don't use any additional low-level graph-cut based methods in training.
The goal is to demonstrate the strength of training with constrained outputs, and how it helps with increasing levels of supervision.

\subsection{Dataset}
\lblsec{dataset}
We evaluate CCNNs for the task of semantic image segmentation on PASCAL VOC dataset~\cite{everingham2014pascal}.
The dataset contains pixel-level labels for $20$ object classes and a separate background class.
For a fair comparison to prior work, we use the similar setup to train all models.
Training is performed on the union of VOC 2012 train set and the larger dataset collected by Hariharan \etal~\cite{BharathICCV2011} summing upto a total of 10,582 training images.
The VOC12 validation set containing a total of 1449 images is kept held-out during ablation studies.
The VGG network architecture used in our algorithm was pre-trained on ILSVRC dataset~\cite{imagenet} for classification task of 1K classes~\cite{simonyan2014very}.

Results are reported in the form of standard intersection over union (IoU) metric, also known as Jaccard Index. 
It is defined per class as the percentage of pixels predicted correctly out of total pixels labeled or classified as that class.
Ablation studies and comparison with baseline methods for both the weak settings are presented in the following subsections.

\subsection{Training from image-level tags}
We start by training our model using just image-level tags.
We obtain these tags from the presence of a class in the pixel-wise ground truth segmentation masks.
The constraints used in this setting are described in Equations~\eqref{eq:suppConstraint},~\eqref{eq:foreConstraint} and~\eqref{eq:backConstraint}.
Since some of the baseline methods report results on the VOC12 validation set, we present the performance on both validation and test set.
Some methods boost their performance by using a Dense CRF model~\cite{kk-eifcc-11} to post process the final output labeling.
To allow for a fair comparison, we present results both with and without a Dense CRF.

\reftbl{weak_fair} compares all contemporary weak segmentation methods.
Our proposed method, CCNN, outperforms all prior methods for weakly labeled semantic segmentation by a significant margin.
MIL-FCN~\cite{pathak2014fully} is an extension of learning based on maximum scoring instance based MIL to multi-class segmentation.
The algorithm proposed by Pinheiro \etal~\cite{pinheiro2014weakly} introduces a soft version of MIL. 
It is trained on $0.7$ million images for 21 classes taken from ILSVRC13, which is $70$ times more data than all other approaches used.
They achieve boost in performance by re-ranking the pixel probabilities with the image-level priors (ILP) i.e. the probability of class to be present in the image.
This suppresses the negative classes and smooths out the predicted segmentation mask.
For the EM-Adapt~\cite{papandreou2015weakly} algorithm, we reproduced the models using their publicly available implementation\footnote{\url{https://bitbucket.org/deeplab/deeplab-public/overview}}.
We apply similar set of constraints on EM-Adapt to make sure it is purely a comparison of the approach.
Note that unconstrained MIL based approach require the final 21-class classifier to be well initialized for reasonable performance.
While our constrained optimization can handle arbitrary random initializations.

We also directly compare our algorithm against the EM-Adapt results as reported in Papandreou \etal~\cite{papandreou2015weakly} for weak segmentation. 
However, their training procedure uses random crops of both the original image and the segmentation mask.
The weak labels are then computed from those random crops. 
This introduces limited information about the spatial location of the weak tags.
Taken to the extreme, a $1\times1$ output crop reduces to full supervision.
We thus present this result in the next subsection on incorporating increasing supervision.

\subsection{Training with additional supervision}
We now consider slightly more supervision than just the image-level tags.
Firstly, we consider the training with tags on random crops of original image, following Papandreou \etal~\cite{papandreou2015weakly}. 
We evaluate our constrained optimization on the EM-Adapt architecture using random crops, and compare to the result obtained from their released caffemodel as shown in \reftbl{weak_additional}. 
Using limited spatial information our algorithm slightly outperforms EM-Adapt, mainly due to the more powerful background constraints.
Note that the difference is not as striking as in the pure weak label setting.
We believe this is due to the fact that the spatial information in combination with the foreground prior emulates the upper bound constraint on background, as a random crop is likely to contain much fewer labels.

\begin{table}[h]
\begin{center}
\resizebox{\linewidth}{!}{%
\begin{tabular}{llcc}
\toprule
\multirow{2}{*}{Method} & \multirow{2}{*}{Training Supervision} & mIoU & mIoU\\
 & & w/o CRF & \\
\midrule
EM-Adapt~\cite{papandreou2015weakly} & Tags w/ random crops & 34.3 & 36.0\\
CCNN & Tags w/ random crops & \textbf{34.4} & \textbf{36.4}\\
\midrule
EM-Adapt~\cite{papandreou2015weakly} & Tags w/ object sizes & -- & --\\
CCNN & Tags w/ object sizes & \textbf{40.5} & \textbf{42.4}\\
\bottomrule
\end{tabular}
}
\end{center}
\vspace{-1em}
\caption{Results using additional supervision during training evaluated on the VOC 2012 validation set.}
\lbltbl{weak_additional}
\end{table}

\begin{figure*}[t]
  \centering
  \begin{subfigure}[b]{0.975\linewidth}
    \centering
    \includegraphics[width=0.24\linewidth]{images/im2_im} %
    \includegraphics[width=0.24\linewidth]{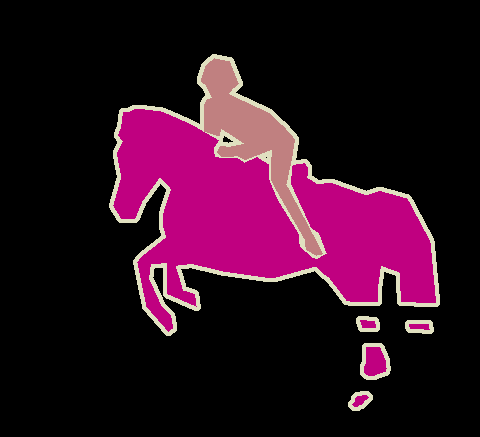}
    \includegraphics[width=0.24\linewidth]{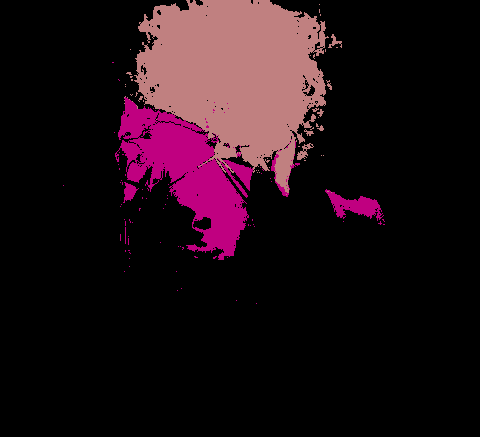}
    \includegraphics[width=0.24\linewidth]{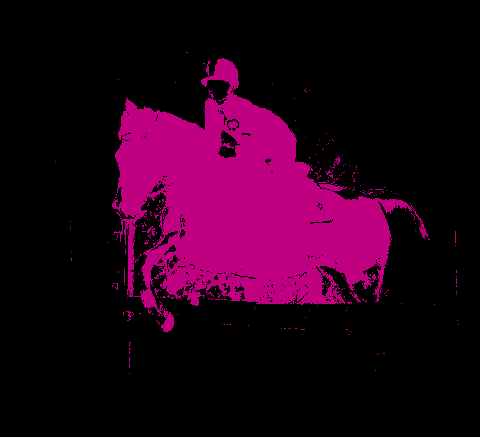}
    \vspace{2pt}
  \end{subfigure}
  \begin{subfigure}[b]{0.975\linewidth}
    \centering
    \includegraphics[width=0.24\linewidth]{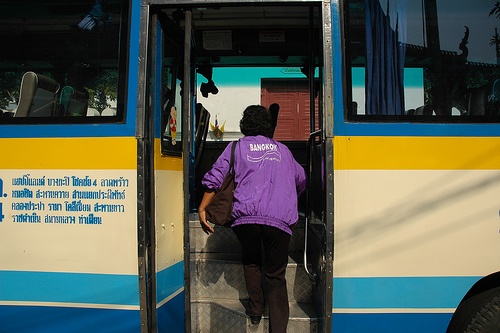}
    \includegraphics[width=0.24\linewidth]{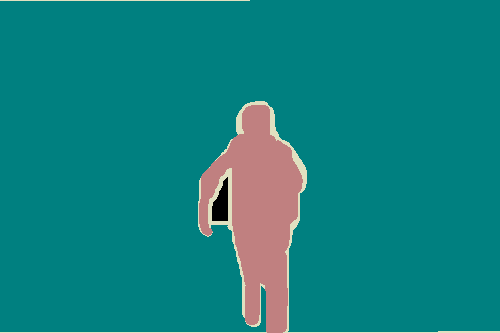}
    \includegraphics[width=0.24\linewidth]{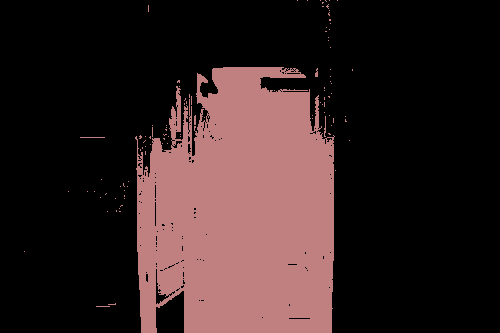}
    \includegraphics[width=0.24\linewidth]{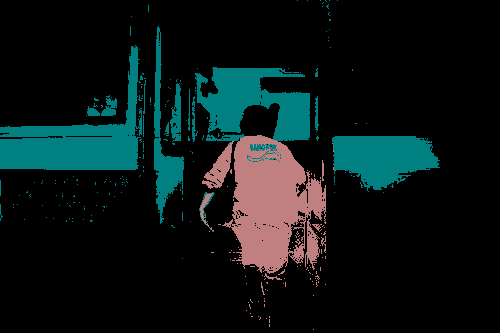}
    \vspace{2pt}
  \end{subfigure}
  \begin{subfigure}[b]{0.975\linewidth}
    \centering
    \includegraphics[width=0.24\linewidth]{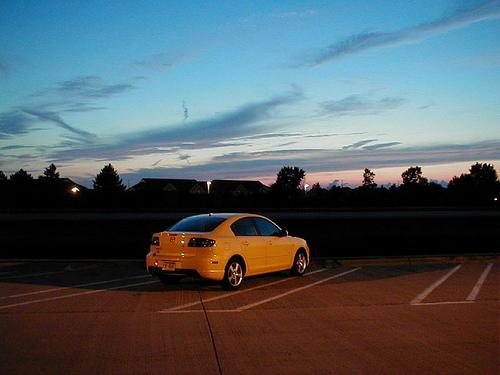}
    \includegraphics[width=0.24\linewidth]{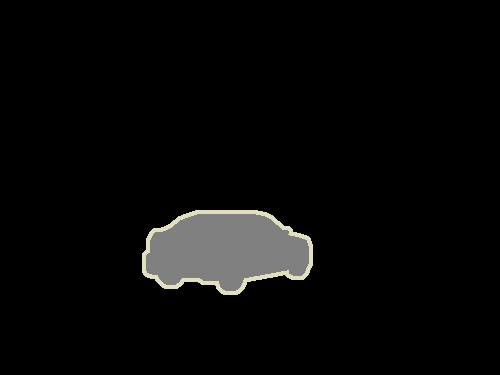}
    \includegraphics[width=0.24\linewidth]{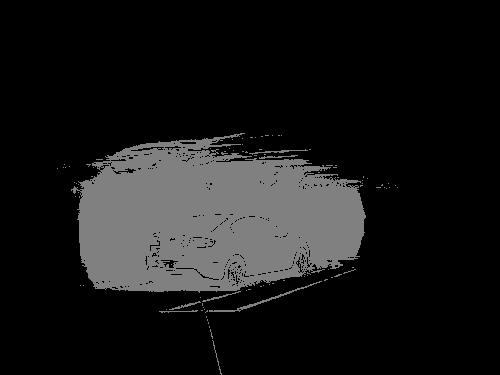}
    \includegraphics[width=0.24\linewidth]{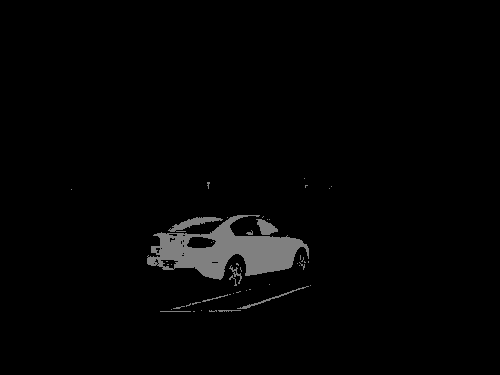}
    \vspace{2pt}
  \end{subfigure}
  \begin{subfigure}[b]{0.975\linewidth}
    \centering
    \includegraphics[width=0.24\linewidth]{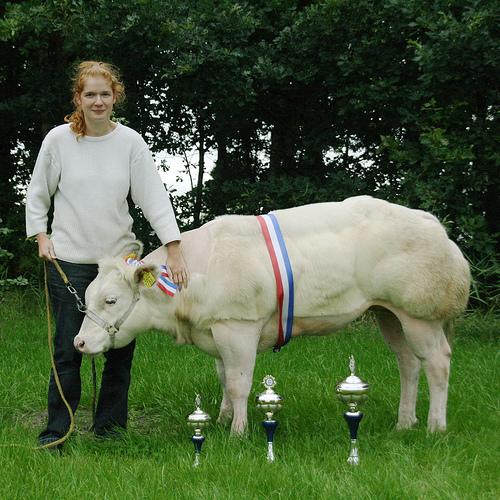}
    \includegraphics[width=0.24\linewidth]{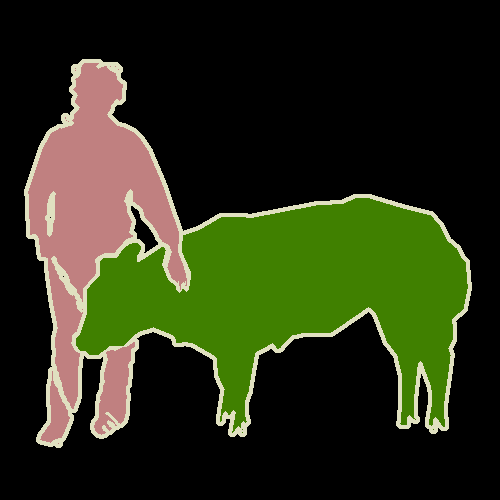}
    \includegraphics[width=0.24\linewidth]{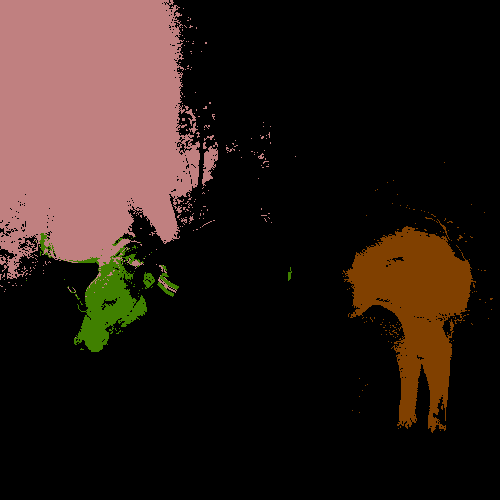}
    \includegraphics[width=0.24\linewidth]{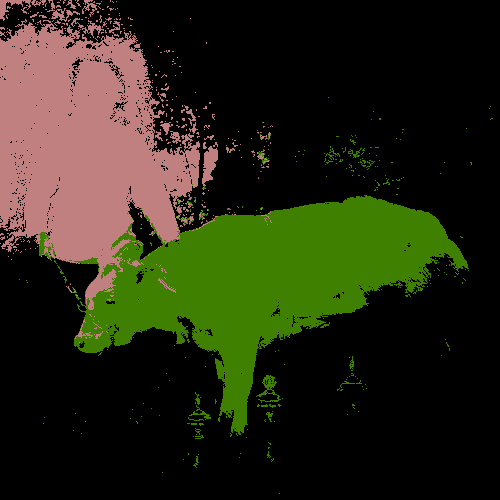}
    \vspace{2pt}
  \end{subfigure}
  \begin{subfigure}[b]{0.24\linewidth}
    \centering
    \caption{Original image}
  \end{subfigure}
  \begin{subfigure}[b]{0.24\linewidth}
    \centering
    \caption{Ground truth}
  \end{subfigure}
  \begin{subfigure}[b]{0.24\linewidth}
    \centering
    \caption{Image tags}
  \end{subfigure}
  \begin{subfigure}[b]{0.24\linewidth}
    \centering
    \caption{Image tags + size}
  \end{subfigure}
  \vspace{-3mm}
  \caption{Qualitative results on the VOC 2012 dataset for different levels of supervision. We show the original image, ground truth, our trained classifier with image level tags and with size constraints. Note that the size constraints localize the objects much better than just image level tags at the cost of missing small objects in few examples.}
  \lblfig{voc_images}
  \vspace{-1mm}
\end{figure*}

The main advantage of CCNN is that there is no restriction of the type of linear constraints that can be used.
To demonstrate this further, we incorporate a simple size constraint.
For each label, we use one additional bit of information: whether a certain class occupies more than $10\%$ of the image or not.
This additional constraint is described in Equation~\eqref{eq:sizeConstraint}.
As shown in \reftbl{weak_additional}, using this one additional bit of information dramatically increases the accuracy.
Unfortunately, EM-Adapt heuristic cannot directly incorporate this more meaningful size constraint.

\reftbl{weak_test} reports our results on PASCAL VOC 2012 test server and compares it to fully supervised approaches.
To better compare with these methods, we further add a result where the CRF parameters are tuned on 100 validation images.
As a final experiment, we gradually add fully supervised images in addition to our weak objective and evaluate the model, i.e., semi-supervised learning. 
The graph is shown in the supplementary material.
Our model makes good use of the additional supervision.

We also evaluate the sensitivity of our model to the parameters of the constraints.
We performed line search along each of the bounds while keeping others fixed.
In general, our method is very insensitive to wide range of constraint bounds due to the presence of slack variables.
The standard deviation in accuracy, averaged over all parameters, is $0.73\%$.
Details are provided in the supplementary material.

Qualitative results are shown in \reffig{voc_images}.

\subsection{Discussion}
We further experimented with bounding box constraints.
We constrain $75\%$ of pixels within a bounding box to take a specific label, while we suppress any labels outside the bounding box.
This additional supervision allows us to boost the IoU accuracy to $54\%$.
This number is competitive with a baseline for which we train a model on all pixels within a bounding box, which gives $52.3\%$~\cite{papandreou2015weakly}.
However it is not yet competitive with more sophisticated systems that use more segmentation information within bounding boxes~\cite{papandreou2015weakly,dai2015boxsup}.
Those systems perform at roughly $58.5-62.0\%$ IoU accuracy.
We believe the key to this performance is a stronger use of the pixel level segmentation information.
In conclusion, we presented CCNN which is a constrained optimization framework to optimize convolutional networks.
The framework is general and can incorporate arbitrary linear constraints.
It naturally integrates into standard Stochastic Gradient Descent, and can easily be used in publicly available frameworks such as Caffe~\cite{caffe}.

We showed that constraints are a natural way to describe the desired output space of a labeling and can reduce the amount of strong supervision CNNs require.

\paragraph{Acknowledgments}
This work was supported by DARPA, AFRL, DoD MURI award N000141110688, NSF awards IIS-1427425 and IIS-1212798, and the Berkeley Vision and Learning Center.

\twocolumn[{\center \Large \bf Supplementary Material:\\
Constrained Convolutional Neural Networks for\\
Weakly Supervised Segmentation
\vspace*{24pt}}]

This document provides detailed derivation for the results used in the paper and additional quantitative experiments to validate the robustness of CCNN algorithm.

In the paper, we optimize constraints on CNN output by introducing latent probability distribution $P(X)$.
Recall the overall main objective as follows:
\begin{align}
\underset{\theta,P}{\text{minimize}} & & &-H_P \underbrace{-\mathbb{E}_{X\sim P}\left[\log Q(X|\theta)\right]}_{H_{P|Q}} \nonumber & \\
\text{subject to} & & & A \vec{P} \geq \vec b, \qquad \sum_X P(X)=1, &\lbleq{KL}
\end{align}
where $H_P=-\sum_{X} P(X)\log P(X)$ is the entropy of latent distribution, $H_{P|Q}$ is the cross entropy and $\vec P$ is the vectorized version of $P(X)$.
In this supplementary material, we will show how to minimize the objective with respect to $P$.
For the complete block coordinate descent minimization algorithm with respect to both $P$ and $\theta$, see the main paper.

Note that the objective function in \eqref{eq:KL} is KL Divergence of network output distribution $Q(X|\theta)$ from $P(X)$, which is convex.
Equation \eqref{eq:KL} is convex optimization problem, since all constraints are linear.
Furthermore, Slaters condition holds as long as the constraints are satisfiable and hence we have strong duality  \cite{boyd2004convex}.
First, we will use this strong duality to show that the minimum of \eqref{eq:KL} is a fully factorized distribution.
We then derive the dual function (Equation \eqref{eq:dual} in the main paper), and finally extend the analysis for the case when objective is relaxed by adding slack variable.

\section*{I. Latent Distribution}
In this section, we show that the latent label distribution $P(X)$ can be modeled as the product of independent marginals without loss of generality.
This is equivalent to showing that the latent distribution that achieves the global optimal value factorizes, while keeping the network parameters $\theta$ fixed.
First, we simplify the cross entropy term in the objective function in~\eqref{eq:KL} as follows:
\begin{align}
H_{P|Q} &= - \mathbb{E}_{X\sim P}\left[\log Q(X|\theta)\right] \nonumber & \\
&= - \mathbb{E}_{X\sim P}\left[\sum_{i=1}^{n} \log q_i(x_i|\theta)\right] \nonumber &
\end{align}
\begin{align}
\quad&= - \sum_{i=1}^{n} \mathbb{E}_{X \sim P}\left[\log q_i(x_i|\theta)\right] \nonumber & \\
&= - \sum_{i=1}^{n} \mathbb{E}_{x_i \sim P}\left[\log q_i(x_i|\theta)\right] \nonumber & \\
&= - \sum_{i=1}^{n} \sum_{l\in\mathcal{L}} P(x_i=l)\log q_i(l|\theta) &\lbleq{simplifiedKL}
\end{align}
We used the linearity of expectation and the fact that $q_i$ is independent of any variable $X_j$ for $j \ne i$ to simplify the objective, as shown above.
Here, $P(x_i=l) = \sum_{X:x_i=l} P(X)$ is the marginal distribution.

Let's now look at the Lagrangian dual function
\begin{align}
\mathcal{L}(P,\lambda,\nu)=&-H_P+H_{P|Q}\notag\\
&+ \lambda^\top(\vec b - A \vec P) + \nu \left(\sum_X P(X) - 1\right)\notag\\
=&-H_P+H_{P|Q}- \sum_{i,l} \lambda^\top A_{i;l} \vec P(x_i=l)\notag\\
&+\lambda^\top\vec b + \nu \left(\sum_X P(X) - 1\right)\notag\\
= &-H_P \underbrace{- \sum_{i=1}^{n} \sum_{l\in\mathcal{L}} P(x_i=l) \left(\log q_i(l|\theta) + A_{i;l}^T\lambda\right)}_{\tilde H_{P|Q}}  \nonumber \\
&+ \lambda^T\vec{b} + \nu \left(\sum_X P(X) - 1\right),\lbleq{bias_ce}
\end{align}
where $\tilde H_{P|Q}$ is a biased cross entropy term and $A_{i;l}$ is the column of $A$ corresponding to $p_i(l)$..
Here we use the fact that the linear constraints are formulated on the marginals to rephrase the dual objective.
We will now show this objective can be rephrased as a KL-divergence between a biased distribution $\tilde Q$ and $P$.

The biased cross entropy term can be rephrased as a cross entropy between a distribution $\tilde Q(X|\theta,\lambda) = \prod_i \tilde q_i(x_i|\theta,\lambda)$, where $\tilde q_i(x_i|\theta,\lambda) = \frac{1}{\tilde Z_i} q_i(x_i|\theta) \exp({A^\top_{i;x_i}\lambda})$ is the biased CNN distribution and $\tilde Z_i$ is a local partition function ensuring $\tilde q_i$ sums to $1$.
This partition function is defined as
$$
 \tilde Z_i = \sum_l \exp\left( \log q_i(l|\theta) + A^\top_{i;l}\lambda \right)
$$
The cross entropy between $P$ and $\tilde Q$ is then defined as

\begin{align}
 H_{P|\tilde Q} &= -\sum_X P(X) \log\tilde Q(X|\theta,\lambda) \notag\\
 &= -\sum_i\!\sum_l\!P(x_i\!=\!l) \log\tilde q_i(l|\theta,\lambda) \notag\\
 &= -\sum_i\!\sum_l\!P(x_i\!=\!l) (\log q_i(l|\theta,\lambda)\!+\!A^\top_{i;l}\lambda\!-\!\log \tilde Z_i) \notag\\
 &= \tilde H_{P|Q} + \sum_i \log \tilde Z_i
\end{align}

This allows us to rephrase \eqref{eq:bias_ce} in terms of a KL-divergence between $P$ and $\tilde Q$
\begin{align}
 &\mathcal{L}(P,\lambda,\nu)\notag\\
 =&-\!H_P+H_{P|\tilde Q} - \sum_i \log \tilde Z_i\!+\!\lambda^T\vec{b}\!+\!\nu\!\left(\!\sum_X P(X) - 1\!\right)\notag\\
 =& D(P||\tilde Q) -C+ \lambda^T\vec{b} + \nu \left(\sum_X P(X) - 1\right)\lbleq{bias_kl},
\end{align}
where $C=\sum_i \log \tilde Z_i$ is a constant that depends on the local partition functions of $\tilde Q$.

The primal objective \eqref{eq:KL} can be phrased as $\min_{P} \max_{\lambda \ge 0, \nu} \mathcal{L}(P,\lambda,\nu)$ which is equivalent to the dual objective $\max_{\lambda \ge 0, \nu} \min_{P} \mathcal{L}(P,\lambda,\nu)$, due to strong duality.

Maximizing the dual objective can be phrased as maximizing a dual function
\begin{align}
\mathcal{L}(\lambda) &=\!\lambda^\top\vec b\!-\!C\!+\!\max_\nu \min_P D(P\|\tilde Q)\!+\!\nu\!\left(\!\sum_X P(X)-1\!\right)\notag\\
&=\!\lambda^\top\vec b\!-\!C\!+\!\underbrace{\min_{P:\sum_X P(X)=1} D(P\|\tilde Q)}_{0}\notag\\
&=\!\lambda^\top\vec b\!-\!\sum_i \log \sum_l \exp\left( \log q_i(l|\theta) + A^\top_{i;l}\lambda \right), \lbleq{dual_l}
\end{align}
where the maximization of $\nu$ can be rephrased as a constraint on $P$ i.e. $\sum_X P(X)=1$.
Maximizing \eqref{eq:dual_l} is equivalent to minimizing the original constraint objective \eqref{eq:KL}.

\paragraph{Factorization}
The KL-divergence $D(P||\tilde Q)$ is minimized at $P = \tilde Q = \prod_i \tilde q_i$, hence the minimal value of $P$ fully factorizes over all variables for any assignment to the dual variables $\lambda$.

\paragraph{Dual function}
Using the definition $q_i(l|\theta) = \frac{1}{Z_i}\exp(f_i(l;\theta))$ we can define the dual function with respect to $f_i$
$$
\mathcal{L}(\lambda)\!=\!\lambda^\top \vec b-\sum_i\!\log\!\sum_l\!\exp(f_i(l;\theta)\!+\!A_{i;l}^\top \lambda) + \underbrace{\sum_i \log Z_i}_\text{const.},
$$
where the log partition function is constant and falls out in the optimization.

\section*{II. Optimizing Constraints with Slack Variable}
The slack relaxed loss function is given by
\begin{align}
\underset{\theta,P,\xi}{\text{minimize}} & & &-H_P \underbrace{-\mathbb{E}_{X\sim P}\left[\log Q(X|\theta)\right]}_{H_{P|Q}} + \beta^\top \xi& \\
\text{subject to} & & & A \vec{P} \geq \vec b-\xi, \qquad \xi \ge 0, \qquad \sum_X P(X)=1 &\notag
\end{align}
For any $\beta \le 0$, a value of $\xi \to \infty$ will minimize the objective and hence invalidate the corresponding constraints.
Thus, for the remainder of the section we assume that $\beta > 0$.
The Lagrangian dual to this loss is defined as
\begin{align}
 \mathcal{L}(P,\lambda,\nu,\gamma) =& -H_P + H_{P|Q}  + \beta^\top \xi + \lambda^\top(\vec b-A\vec P-\xi)\notag\\
 &+\nu\left( \sum_X P(X)-1 \right) - \gamma^\top \xi. \lbleq{dual_slack}
\end{align}
We know that the dual variable $\gamma \ge 0$ is strictly non-negative, as well as
\begin{equation}
\frac{\partial}{\partial \xi} \mathcal{L}(P,\lambda,\nu,\gamma) = \beta-\lambda-\gamma = 0. \lbleq{grad}
\end{equation}
This leads to the following constraint on $\lambda$:
$$
 0 \le \gamma = \beta-\lambda.
$$
Hence the slack weight forms an upper bound on $\lambda \le \beta$.
Substituting \eqref{eq:grad} into \eqref{eq:dual_slack} reduces the dual objective to the non-slack objective in \eqref{eq:bias_ce}, and the rest of the derivation is equivalent.

\section*{III. Ablation Study for Parameter Selection}
In this section, we present results to analyze the sensitivity of our approach with respect to the constraint parameters i.e. the upper and lower bounds.
We performed line search along each of the bounds while keeping others fixed.
The method is quite robust with a standard deviation of $0.73\%$ in accuracy, averaged over all parameters, as shown in~\reftbl{ablation}.
These experiments are performed in the setting where image-level tag and 1-bit size supervision is available during training, as discussed in the paper.
We attribute this robustness to the slack variables that are learned per constraint per image.

\begin{table}[h!]
\begin{center}
\resizebox{\linewidth}{!}{%
\begin{tabular}{cccc}
\toprule
Fgnd lower & Bgnd lower & Bgnd upper & mIoU \\
$a_l$ & $a_0$ & $b_0$ & w/o CRF \\
\midrule
\textbf{0.1} & \textbf{0.2} & \textbf{0.7} & \textbf{40.5} \\
\midrule
0.2 & 0.2 & 0.7 & 40.6 \\
0.3 & 0.2 & 0.7 & 40.6 \\
0.4 & 0.2 & 0.7 & 39.6 \\
\midrule
0.1 & 0.1 & 0.7 & 40.5 \\
0.1 & 0.3 & 0.7 & 40.4 \\
0.1 & 0.4 & 0.7 & 40.5 \\
0.1 & 0.5 & 0.7 & 40.4 \\
\midrule
0.1 & 0.2 & 0.5 & 36.6 \\
0.1 & 0.2 & 0.6 & 38.9 \\
0.1 & 0.2 & 0.8 & 39.7 \\
\bottomrule
\end{tabular}
}
\end{center}
\vspace{-1em}
\caption{Ablation study for sensitivity analysis of the CCNN optimization with respect to the chosen parameters. The paramters mentioned here are defined in Equations~\eqref{eq:foreConstraint} and \eqref{eq:backConstraint} in the main paper. Parameter values used in all other experiments are shown in bold.}
\lbltbl{ablation}
\end{table}

\section*{IV. Ablation Study in Semi-Supervised Setting}
In this section, we experiment by incorporating fully supervised images in addition to our weak objective. 
The accuracy curve is depicted in \reffig{semi}.

\begin{figure}[h]
\begin{center}
\includegraphics[width=\linewidth]{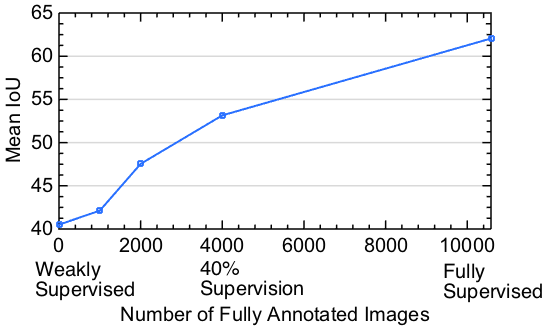}
\end{center}
\vspace{-1em}
\caption{Ablation study with varying amount of fully supervised images.
Our model makes good use of the additional supervision.}
\lblfig{semi}
\end{figure}

\newpage
{\small
\bibliographystyle{ieee}
\bibliography{main}
}

\end{document}